\documentclass[10pt,twocolumn]{article}

\usepackage[a4paper,margin=2cm,columnsep=0.6cm]{geometry}
\usepackage{titlesec}
\usepackage{authblk}

\usepackage[T1]{fontenc}
\usepackage[utf8]{inputenc}
\usepackage{lmodern}
\usepackage{microtype}

\usepackage{amsmath,amssymb,amsfonts}

\usepackage{booktabs}
\usepackage{multirow}
\usepackage{graphicx}
\usepackage{float}
\usepackage{caption}
\usepackage{subcaption}
\usepackage{array}

\usepackage{algorithm}
\usepackage{algorithmic}

\usepackage{listings}
\usepackage[hidelinks,colorlinks=true,citecolor=blue!60!black,linkcolor=blue!60!black,urlcolor=blue!60!black]{hyperref}
\usepackage[numbers,sort&compress]{natbib}

\hypersetup{
  pdftitle={StateSMix: Online Lossless Compression via Mamba State Space Models and Sparse N-gram Context Mixing},
  pdfauthor={Roberto Tacconelli},
  pdfsubject={Neural Lossless Data Compression},
  pdfkeywords={lossless compression, state space models, Mamba, arithmetic coding, n-gram, online learning},
  pdfusetitle=false,
}

\usepackage{xcolor}
\usepackage{enumitem}
\setlist{nosep,leftmargin=*}

\titleformat{\section}{\large\bfseries}{\thesection.}{0.5em}{}
\titleformat{\subsection}{\normalsize\bfseries}{\thesubsection}{0.5em}{}
\titleformat{\subsubsection}{\normalsize\itshape}{\thesubsubsection}{0.5em}{}

\newcommand{\ssmcomp}{\textsc{StateSMix}}
\newcommand{\bpb}{\text{bpb}}
\newcommand{\RR}{\mathbb{R}}

\title{\textbf{StateSMix: Online Lossless Compression via\\
Mamba State Space Models and\\Sparse N-gram Context Mixing}}

\author[1]{Roberto Tacconelli}
\affil[1]{Independent Researcher\\
\texttt{tacconelli.rob@gmail.com}}

\date{}

\begin{document}
\maketitle
\thispagestyle{empty}

\begin{abstract}
We present \ssmcomp{}, a lossless compressor that couples a lightweight
Mamba-style State Space Model (SSM) trained entirely \emph{online} during
compression with a suite of sparse n-gram context models and range
arithmetic coding.
Unlike LLM-based compressors relying on hundreds of millions of
pre-trained parameters stored externally, \ssmcomp{} encodes all model
knowledge implicitly: the model is initialised from scratch and trained
token-by-token on the file being compressed, making the output fully
self-contained.
The SSM ($D_M=32$, $N_L=2$, ${\sim}120$K active parameters per file)
provides a continuously-updated global probability estimate, while
sparse n-gram tables (bigram through 32-gram, $2^{24}=16\text{M}$ slots
each) add exact local-pattern memorisation via a
\emph{softmax-invariant logit-bias} mechanism that touches only
non-zero-count tokens.  Long-range context tables (16-gram and 32-gram)
extend the n-gram model to capture repeated multi-token sequences
beyond the SSM's recurrent memory horizon.
On the standard enwik8 benchmark, \ssmcomp{} achieves
\textbf{2.161\,\bpb{}} on the 10\,MB excerpt,
\textbf{beating xz~$-$9e by 0.7\%}, and 2.130\,\bpb{} on the full
100\,MB corpus.
Ablation experiments on enwik8\textsubscript{3M} demonstrate that the
SSM is the dominant contributor, accounting for a $46.6\%$ size reduction
over a frequency-count baseline and beating xz even without any n-gram
component, while n-gram models without the SSM achieve only $16.1\%$
reduction.
With OpenMP parallelisation of the training loop, the system achieves
${\sim}2{,}000$\,tok/s on 4 cores with AVX2 SIMD; no GPU or
pre-trained weights are required.
\end{abstract}


\medskip
\noindent\textbf{Code:} \url{https://github.com/robtacconelli/StateSMix}

\section{Introduction}
\label{sec:intro}

Shannon's source coding theorem~\citep{shannon1948} establishes that the
minimum code length for a symbol drawn from source $P$ is
$-\log_2 P(\text{symbol})$ bits.  Compression is therefore equivalent
to prediction: a model that assigns high probability to the next symbol
enables an arithmetic coder to represent it efficiently~\citep{rissanen1979}.
This equivalence has driven a long progression of compressors, from
LZ77~\citep{ziv1977} and LZMA through PPM~\citep{cleary1984}, the
PAQ/CMIX context-mixing family~\citep{mahoney2005,cmix2024}, and
recently neural approaches using LSTM or Transformer language
models~\citep{bellard2021,deletang2024,huang2024finezip}.

A key tension in neural compression is between \emph{model quality}
(how well the model predicts the specific input) and \emph{model cost}
(parameters that must be transmitted with the archive, or inference
time per token).  LLM-based compressors such as ts\_zip~\citep{bellard2023}
and FineZip~\citep{huang2024finezip} achieve impressive ratios but require
hundreds of megabytes of external weights and GPU inference, making them
impractical for general use.

We explore a complementary regime: \emph{fully online} neural
compression, where the model is trained from random initialisation
on the file being compressed.  No pre-trained weights are needed;
all model knowledge is implicitly encoded in the compressed bitstream.
This paradigm was pioneered by NNCP~\citep{bellard2021} using
Transformer-XL but incurs prohibitive per-token training cost.  We
instead use a Mamba-style State Space Model (SSM)~\citep{gu2023mamba}
with DM=32 and NL=2, which affords linear-time inference, a compact
recurrent state, and fast backpropagation---all affordable in pure C
with AVX2 SIMD at ${\sim}1{,}300$\,tok/s without any GPU.

The core contributions of \ssmcomp{} are:
\begin{enumerate}
\item \textbf{Online Mamba compression.}  A two-layer Mamba SSM serves
  as the primary online predictor, trained by Adam gradient descent
  after every 32-token chunk, requiring no pre-training, no external
  weights, and no GPU.

\item \textbf{Softmax-invariant logit bias.}  We derive a sparse update
  formula for adding n-gram evidence that exploits softmax translation
  invariance, touching only tokens with non-zero counts and making
  high-order (2--8-gram) context models memory- and compute-efficient.

\item \textbf{Entropy-adaptive mixing.}  N-gram bias magnitude is scaled
  by the SSM's predictive entropy, so n-grams contribute more when the
  SSM is uncertain and retreat when it is confident.

\item \textbf{Compact vocabulary remapping.}  Only tokens present in
  the current file are modelled, reducing the effective vocabulary
  from 49{,}152 to 18K--44K and cutting head-projection cost by
  10--30\%.

\item \textbf{Linear-probing collision resolution.}  Open-addressed
  n-gram hash tables use probe depth 8, recovering contexts lost under
  simple replacement, critical at the load factors seen in large files.
\end{enumerate}

Empirically, the SSM alone beats xz on enwik8\textsubscript{3M}
(840\,KB vs.\ 852\,KB), while the full system achieves a further
3.7\% reduction.  The crossover to xz superiority occurs around 30\,MB,
driven by LZMA's ability to exploit long-distance repetitions beyond the
reach of fixed-horizon n-gram tables.

The remainder of the paper is structured as follows.
Section~\ref{sec:related} surveys related work.
Section~\ref{sec:background} introduces SSMs and the online learning
framework.
Section~\ref{sec:method} describes the \ssmcomp{} architecture in
detail.
Section~\ref{sec:theory} provides theoretical analysis.
Section~\ref{sec:experiments} presents experiments and ablation.
Section~\ref{sec:discussion} discusses implications and
Section~\ref{sec:conclusion} concludes.

\section{Related Work}
\label{sec:related}

\subsection{Classical Lossless Compression}

Dictionary-based methods --- LZ77~\citep{ziv1977}, gzip/DEFLATE,
LZMA/xz, Zstandard --- exploit byte repetitions within a sliding
window, achieving ${\sim}2.0$\,\bpb{} on English text.
Huffman coding~\citep{huffman1952} assigns variable-length codes by
symbol frequency; arithmetic coding~\citep{witten1987} removes the
integer-bit constraint, approaching entropy to within a fraction of a
bit.  Among classical tools, \texttt{xz}~$-$9e (LZMA2) achieves
${\sim}1.99$\,\bpb{} on enwik8 and serves as our primary baseline.

\subsection{Context Mixing and Adaptive Statistical Models}

PPM~\citep{cleary1984} uses adaptive variable-order context modelling
with arithmetic coding, achieving ${\sim}2.0$\,\bpb{} on English
text.  The PAQ family~\citep{mahoney2005} extends this with context
mixing: blending hundreds of specialised models via neural networks
at the bit level.  PAQ8px achieves ${\sim}1.27$\,\bpb{} on enwik8
(at the $-$12L setting) but requires extreme compute and memory.
CMIX~\citep{cmix2024} incorporates LSTM networks alongside thousands
of context models, reaching ${\sim}1.17$\,\bpb{} on enwik8 at
$0.5$--$5$\,KB/s with 16--64\,GB RAM.

NNCP~\citep{bellard2021} uses a Transformer-XL trained \emph{online}
during compression, achieving ${\sim}1.19$\,\bpb{} on enwik8 but
storing model weights (${\sim}10$\,MB) in the compressed output.

\subsection{LLM-Based Compression}

\citet{deletang2024} showed that Chinchilla~70B achieves
0.664\,\bpb{} on enwik9 via arithmetic coding.  FineZip~\citep{huang2024finezip}
uses LLaMA-3-8B with LoRA fine-tuning, achieving 1.024\,\bpb{} on
enwik8.  Bellard's ts\_zip~\citep{bellard2023} uses RWKV-169M with
8-bit quantisation, achieving ${\sim}1.11$\,\bpb{} on enwik8.
DeepZip~\citep{goyal2020} combined recurrent networks with arithmetic
coding for general-purpose compression.

All LLM-based compressors require the model weights to be transmitted
or pre-shared; the compressed output is not self-contained.  Our
approach is the opposite: the model is trained online and transmitted
implicitly.

\subsection{State Space Models in Sequence Modelling}

S4~\citep{gu2021s4} demonstrated that structured SSMs with HiPPO
initialisation capture long-range dependencies efficiently.
Mamba~\citep{gu2023mamba} introduced input-dependent selection into
the SSM, making $\mathbf{B}$, $\mathbf{C}$, and $\Delta t$ functions
of the current input.  Mamba-2~\citep{dao2024mamba2} further refined
the selective SSM formulation.  To our knowledge, we are the first to
apply a Mamba-style online-trained SSM to lossless compression.

Table~\ref{tab:landscape} positions \ssmcomp{} in the broader
compression landscape.

\begin{table*}[t]
\centering
\caption{Lossless compressor landscape on enwik8 (100\,MB).
  $\dagger$~Pre-trained model weights excluded from compressed size;
  must be pre-shared.
  $\ddagger$~Model weights stored in compressed output.
  ``Online'' = model trained from scratch on the compressed file;
  no pre-trained weights required.
}
\label{tab:landscape}
\small
\begin{tabular}{@{}llrrrccl@{}}
\toprule
\textbf{System} & \textbf{Model type} & \textbf{Params} & \textbf{\bpb{}} & \textbf{Speed} & \textbf{GPU} & \textbf{Self-cont.} & \textbf{Category} \\
\midrule
gzip (DEFLATE)      & ---                & ---     & 2.916 & GB/s       & \textemdash & \checkmark & dictionary \\
xz (LZMA2)          & ---                & ---     & 1.989 & 10\,MB/s   & \textemdash & \checkmark & dictionary \\
bzip2               & ---                & ---     & 2.321 & 50\,MB/s   & \textemdash & \checkmark & BWT+Huffman \\
PAQ8px              & context mixing     & ---     & ${\sim}1.27$ & $<$1\,KB/s & \textemdash & \checkmark & context mix \\
CMIX v21            & LSTM + mixing      & ${\sim}50$M & ${\sim}1.17$ & $<$1\,KB/s & optional & \checkmark & context mix \\
\midrule
NNCP v3$^\ddagger$  & Transformer-XL     & online  & ${\sim}1.19$ & ${\sim}$1\,KB/s & optional & \checkmark & online neural \\
\midrule
Del\'{e}tang et al.$^\dagger$ & Chinchilla & 70B & ${\approx}1.6$ & --- & \checkmark & \textemdash & LLM \\
FineZip$^\dagger$   & LLaMA-3-8B         & 8B     & 1.024   & --- & \checkmark & \textemdash & LLM \\
ts\_zip$^\dagger$   & RWKV-169M          & 169M   & ${\sim}1.11$ & --- & optional & \textemdash & LLM \\
Nacrith~\citep{tacconelli2026}$^\dagger$ & SmolLM2 + mixing & 135M & 0.939 & --- & optional & \textemdash & LLM \\
\midrule
\textbf{\ssmcomp{} (ours)} & \textbf{Mamba SSM} & \textbf{online} & \textbf{2.130} & \textbf{${\sim}$700\,B/s} & \textbf{\textemdash} & \textbf{\checkmark} & \textbf{online neural} \\
\bottomrule
\end{tabular}
\end{table*}

\section{Background}
\label{sec:background}

\subsection{Arithmetic Coding and Prediction}

Arithmetic coding~\citep{witten1987} encodes a sequence
$(t_1,\ldots,t_N)$ as a single real number in $[0,1)$ by
progressively narrowing an interval according to each token's
probability.  Given a predictor $q(\cdot|\text{context})$, the
expected code length is $\sum_i -\log_2 q(t_i\,|\,t_{<i})$ bits,
equalling the cross-entropy between the true source and the model.
For lossless reconstruction, the decoder must generate the \emph{same}
sequence of distributions $q_1, q_2, \ldots$; in \ssmcomp{} this is
guaranteed because both encoder and decoder update the model with the
true token after each step, maintaining identical state.

\subsection{State Space Models}

The continuous-time SSM~\citep{kalman1960} is defined by:
\begin{equation}
  \mathbf{h}'(t) = \mathbf{A}\,\mathbf{h}(t) + \mathbf{B}\,\mathbf{x}(t),
  \qquad
  \mathbf{y}(t) = \mathbf{C}\,\mathbf{h}(t) + \mathbf{D}\,\mathbf{x}(t)
\end{equation}
where $\mathbf{h}(t)\in\RR^{D_S}$ is the recurrent hidden state.
Discretising with step $\Delta$:
\begin{align}
  \mathbf{h}_i &= e^{\Delta\mathbf{A}}\mathbf{h}_{i-1}
              + \Delta\mathbf{B}\,\mathbf{x}_i,
  \quad
  \mathbf{y}_i = \mathbf{C}\mathbf{h}_i + \mathbf{D}\mathbf{x}_i.
\end{align}
S4~\citep{gu2021s4} showed that with HiPPO initialisation for
$\mathbf{A}$, the discrete SSM captures long-range dependencies
efficiently.  Mamba~\citep{gu2023mamba} introduced \emph{selectivity}:
$\mathbf{B}$, $\mathbf{C}$, and $\Delta$ become input-dependent,
giving the model per-token control over state retention.  This enables
Mamba to forget irrelevant context and focus on salient tokens, while
retaining $O(N)$ inference complexity---ideal for online token-by-token
processing.

\subsection{Online Learning for Compression}

In online prediction, a learner outputs $q_i$ before seeing $t_i$,
then updates.  The \emph{regret} after $N$ steps is
$R_N = \sum_i -\log q_i(t_i) - \min_q \sum_i -\log q(t_i)$.
A good online learner minimises $R_N$, compressing as if it had known
the best fixed model in advance.  For neural compressors, stochastic
gradient descent on the online cross-entropy loss (updating after each
chunk) provides a practical online learning algorithm with strong
empirical performance.

\section{Method}
\label{sec:method}

\subsection{System Overview}

\ssmcomp{} operates in four stages:
(1) BPE tokenisation of the raw input,
(2) compact vocabulary remapping,
(3) online predict-encode-update loop, and
(4) file serialisation.
Decompression is the mirror image: the decoder runs the same
predict-update loop, recovering each token from the arithmetic decoder.
Algorithm~\ref{alg:compress} outlines the pipeline.

\begin{algorithm}[H]
\caption{\ssmcomp{} compression}
\label{alg:compress}
\begin{algorithmic}[1]
\REQUIRE raw bytes $\mathbf{s}$
\STATE $(t_1,\ldots,t_N) \leftarrow \mathrm{BPE}(\mathbf{s})$
\STATE $v_e,\,\phi \leftarrow \mathrm{BuildVocabMap}(\mathbf{t})$; remap $\mathbf{t}$
\STATE Init SSM params $\theta$, n-gram tables, ArithEncoder
\FOR{$i = 1$ to $N$}
  \STATE $\boldsymbol{\ell} \leftarrow \mathrm{SSMForward}(\theta, t_{i-1})$
  \STATE $\boldsymbol{\ell} \mathrel{+}= \mathrm{NgramBias}(t_{<i})$ \quad [Eq.~\ref{eq:bias}]
  \STATE $p \leftarrow \mathrm{softmax}(\boldsymbol{\ell})$; encode $t_i$
  \IF{$i \bmod 32 = 0$}
    \STATE $\mathrm{SSMTrain}(\theta, t_{i-31:i})$ \quad [Adam, $n_\mathrm{iter}$ steps]
  \ENDIF
  \STATE $\mathrm{NgramUpdate}(t_{<i},\, t_i)$
\ENDFOR
\RETURN ArithEncoder.finish()
\end{algorithmic}
\end{algorithm}

\subsection{BPE Tokenisation and Compact Vocabulary}
\label{sec:vocab}

Raw bytes are tokenised using a Byte Pair Encoding (BPE) tokeniser
with the GPT-NeoX vocabulary ($V=49{,}152$ types).  BPE converts
variable-length byte sequences into discrete token IDs; English
Wikipedia text tokenises at ${\sim}3.3$--$3.5$ bytes/token.

Since only $v_e \ll V$ token types appear in any given file, \ssmcomp{}
builds a bijective compact remapping $\phi: [0,v_e)\to[0,V)$ from
compact IDs to vocabulary IDs.  All SSM embeddings, head weights, and
n-gram counts are allocated and computed only over $v_e$ tokens,
reducing per-token operations from $O(V\cdot D_M)$ to
$O(v_e\cdot D_M)$.  For enwik8 (100\,MB), $v_e=44{,}298$---a
${\sim}10\%$ reduction; for 1\,MB excerpts, $v_e=18{,}058$---a $63\%$
reduction.

The mapping $\phi$ is Rice-coded~\citep{rice1979} and stored in the
compressed header.  For $v_e=44{,}298$, it requires ${\sim}12$\,KB
versus $4V=197$\,KB for an uncompressed lookup table.

\subsection{Mamba SSM Architecture}
\label{sec:ssm}

The predictor consists of $N_L=2$ Mamba layers, a final layer
normalisation, and a linear language model head.

\subsubsection{Embedding and Head Projection}

Each token $t$ maps to an embedding $\mathbf{e}_t\in\RR^{D_M}$
(a learnable matrix $\mathbf{E}\in\RR^{v_e\times D_M}$ indexed by
compact ID).  After the final layer, the normalised hidden state
$\mathbf{x}_\text{fin}\in\RR^{D_M}$ is projected to logits:
\begin{equation}
  \ell_j = \mathbf{w}_j^\top\!\mathbf{x}_\text{fin},
  \quad j=0,\ldots,v_e-1
\end{equation}
where $\mathbf{W}\in\RR^{v_e\times D_M}$ is the head matrix.  At
$v_e=44{,}298$ and $D_M=32$, this projection (${\sim}1.4$M MADs) is
the per-token bottleneck.

\subsubsection{Mamba Layer}

Each layer processes input $\mathbf{x}\in\RR^{D_M}$:

\smallskip
\noindent\textbf{Layer normalisation.}
$\hat{\mathbf{x}} = \mathrm{LN}_{\mathbf{g},\mathbf{b}}(\mathbf{x})$.

\smallskip
\noindent\textbf{Input projection.}
A weight matrix $\mathbf{W}_\text{in}\in\RR^{D_M\times 2D_I}$ maps
$\hat{\mathbf{x}}$ to two halves of size $D_I=2D_M=64$:
SSM branch $\mathbf{z}_s$ and gate branch $\mathbf{z}_g$.

\smallskip
\noindent\textbf{Depthwise convolution.}
$\tilde{\mathbf{z}}_s = \mathrm{SiLU}(\mathrm{DWConv}_{D_C=4}(\mathbf{z}_s))$.
The causal convolution buffer $\mathrm{cb}\in\RR^{D_I\times(D_C-1)}$
is maintained across tokens.

\smallskip
\noindent\textbf{SSM parameter projection.}
$\mathbf{W}_{xp}\in\RR^{D_I\times(2D_S+1)}$ maps
$\tilde{\mathbf{z}}_s$ to $[\mathbf{B};\mathbf{C};\delta]$
with $D_S=16$:
\begin{gather}
  [\mathbf{B};\mathbf{C};\delta]
  = \tilde{\mathbf{z}}_s\,\mathbf{W}_{xp},\notag\\
  \boldsymbol{\Delta} = \mathrm{softplus}(\delta\cdot\mathbf{w}_\Delta+\mathbf{b}_\Delta)\in\RR^{D_I}.
\end{gather}

\smallskip
\noindent\textbf{Selective SSM recurrence.}
For each channel $i$ and state index $j$:
\begin{align}
  \bar{A}_{i,j} &= \exp(\Delta_i\cdot A_{i,j}), \label{eq:dA}\\
  h_{i,j}       &\leftarrow \bar{A}_{i,j}\,h_{i,j}
                 + \Delta_i\,B_j\,\tilde{z}_{s,i}, \label{eq:hup}\\
  y_i           &= \textstyle\sum_j h_{i,j}\,C_j + D_i\,\tilde{z}_{s,i}.
\end{align}
$A_{i,j}=-\exp(A^\text{log}_{i,j})<0$ ensures stability.
The state $\mathbf{h}\in\RR^{D_I\times D_S}$ is carried across tokens.

\smallskip
\noindent\textbf{Gating and output projection.}
$\mathbf{o} = \mathbf{y}\odot\mathrm{SiLU}(\mathbf{z}_g)$;
$\mathbf{x}' = \mathbf{o}\,\mathbf{W}_\text{out} + \mathbf{x}$
(residual connection back to $\RR^{D_M}$).

The full recurrent state is
$\mathbf{H}\in\RR^{N_L\times D_I\times D_S}
= 2\times64\times16 = 2{,}048$ floats per token position,
plus $N_L\times D_I\times(D_C-1)=384$ floats for convolution buffers.

\subsubsection{Initialisation}

Weights are drawn from $\mathcal{N}(0,0.02^2)$; bias terms from
$\mathcal{N}(0,(0.1)^2)$.
$A^\text{log}_{i,j}$ is initialised as $\log(j+1)$, giving a geometric
spread of decay rates: slow ($j=D_S-1$) to fast ($j=0$), encouraging
specialisation across time scales.
$\mathbf{D}$ is initialised to $\mathbf{1}$ (identity skip).

\subsubsection{Parameter Count}

With $v_e$ effective tokens the model has
$19{,}776 + 2\,v_e\,D_M$ parameters (fixed architecture weights plus
embedding and head), as detailed in Table~\ref{tab:params}.

\begin{table}[h]
\centering
\small
\caption{Parameter count (DM=32, DS=16, DI=64, NL=2).}
\label{tab:params}
\begin{tabular}{@{}lrr@{}}
\toprule
\textbf{Component} & \textbf{Floats} & \textbf{KB} \\
\midrule
LN + in-proj (per layer)       & 4,160 & 16.3 \\
Conv + xp-proj (per layer)     & 2,336 & 9.1  \\
A-log, D, out-proj (per layer) & 3,360 & 13.1 \\
$\times 2$ layers              & 19,712 & 77.0 \\
Final LN                       & 64    & 0.25 \\
\midrule
Embedding $v_e\times32$        & $32\,v_e$ & variable \\
Head $v_e\times32$             & $32\,v_e$ & variable \\
\midrule
\textbf{Total (enwik8)} & \textbf{2,851,888} & \textbf{10,937} \\
\textbf{Total (1\,MB)} & \textbf{1,175,488} & \textbf{4,592} \\
\bottomrule
\end{tabular}
\end{table}

\subsection{Online Training}
\label{sec:training}

Training proceeds simultaneously with encoding.  Tokens are buffered
in chunks of $C=32$; after each chunk, the parameters are updated
for $n_\mathrm{iter}$ Adam steps on the chunk's cross-entropy loss
with label smoothing $\varepsilon=0.12$:
\begin{equation}
  \mathcal{L} = \sum_{i=1}^{C-1}
  \Bigl[(1-\varepsilon)\,\mathrm{CE}(\mathbf{p}_i, t_{i+1})
        + \varepsilon\,H_u\Bigr]
\end{equation}
where $H_u$ is the cross-entropy with the uniform distribution.
Gradients are computed by exact backpropagation through each chunk
with SSM state $\mathbf{H}$ detached at chunk boundaries (truncated BPTT).

A warm-up schedule applies more iterations to early chunks, bootstrapping
the model quickly before the n-gram tables have accumulated enough
observations:
\begin{equation}
n_\mathrm{iter} = \begin{cases}
  8 & \text{chunks } 1\text{--}10, \\
  4 & \text{chunks } 11\text{--}30, \\
  2 & \text{chunks } 31+.
\end{cases}
\end{equation}

Adam hyperparameters: $\eta=0.002$, $\beta_1=0.9$, $\beta_2=0.999$,
$\hat\varepsilon=10^{-8}$, gradient clipping at $G_\text{clip}=5.0$.
The update is applied to all fixed architecture weights plus the
$v_e$-slice of embedding and head matrices (vocab-adaptive).

\subsection{Sparse N-gram Logit Bias}
\label{sec:ngram}

\subsubsection{Softmax Invariance and the Sparse Bias Formula}

Softmax is invariant to additive constants:
$\mathrm{softmax}(\boldsymbol{\ell} + c\,\mathbf{1}) = \mathrm{softmax}(\boldsymbol{\ell})$.
Therefore, when incorporating n-gram evidence into the SSM logit vector,
only tokens with non-zero counts need to be updated.  Define the
n-gram logit delta for context-conditioned counts $\{c_j\}$:
\begin{equation}
  \delta_j = \lambda\,\log\!\Bigl(1 + \frac{c_j}{\alpha}\Bigr),
  \quad c_j > 0
  \label{eq:bias}
\end{equation}
Unseen tokens ($c_j=0$) receive $\delta_j=0$.  Adding $\boldsymbol{\delta}$
to the SSM logit is equivalent to multiplying the corresponding
token probabilities by $e^{\delta_j}/Z_\delta$ (where $Z_\delta$ is
the renormalisation factor from softmax), a principled Bayesian
likelihood update of the SSM prior with evidence proportional to the
smoothed n-gram count.

This formulation makes sparse n-gram storage both memory-efficient
(no dense probability vector needed) and compute-efficient (only
$O(\text{fan-out})$ operations per token, where fan-out is typically
5--30 for natural language).

\subsubsection{Hash Tables and Linear Probing}

For each n-gram order $k\in\{2,\ldots,8,16,32\}$, a hash table of
$2^{24}=16\text{M}$ slots stores per-context sparse count arrays
(Table~\ref{tab:ngrams}).  The \emph{context key} is:
\begin{itemize}
  \item Exactly-packed 64-bit integer for orders 2--4 (16 bits per
    token; $v_e < 2^{16}$).
  \item Murmur-style mix64 hash of a polynomial rolling hash for
    orders 5--8 and the long-range orders (16, 32):
    $k = k \cdot 104729 + (\text{token ID})$,
    then $k \xrightarrow{\text{mix64}} k'$.
\end{itemize}
Collision resolution uses \emph{open addressing with linear probing
depth~8}: on insertion, up to 8 consecutive slots are examined
before the entry is discarded.  The lookup terminates at the first
empty slot or the matching key.  This significantly reduces wasted
slots: at 30\% load, the expected additional probes per lookup is
${\approx}0.21$ (vs.\ total loss under simple replacement).

Each slot stores a full 64-bit context key (not the hash) for
collision rejection, plus a dynamically-grown sparse count array
(uint16 token IDs and counts, typical capacity 4--32 entries).

The bigram ($k=1$) uses a direct array indexed by previous token
(size $v_e$), eliminating hash collisions entirely.

\begin{table}[h]
\centering
\small
\caption{N-gram model parameters ($\lambda$, $\alpha$ as in Eq.~\ref{eq:bias}).}
\label{tab:ngrams}
\begin{tabular}{@{}lcccc@{}}
\toprule
\textbf{Order} & \textbf{Ctx} & \textbf{Slots} & $\boldsymbol{\lambda}$ & $\boldsymbol{\alpha}$ \\
\midrule
bigram  & 1 & $v_e$ & 0.15 & 0.10 \\
trigram & 2 & $2^{24}$ & 0.10 & 0.05 \\
4-gram  & 3 & $2^{24}$ & 0.08 & 0.03 \\
5-gram  & 4 & $2^{24}$ & 0.06 & 0.02 \\
6-gram  & 5 & $2^{24}$ & 0.05 & 0.015 \\
7-gram  & 6 & $2^{24}$ & 0.04 & 0.010 \\
8-gram  & 7 & $2^{24}$ & 0.03 & 0.008 \\
\midrule
16-gram & 15 & $2^{24}$ & 0.50 & 0.001 \\
32-gram & 31 & $2^{24}$ & 1.00 & 0.001 \\
\bottomrule
\end{tabular}
\end{table}

\subsubsection{Long-Range Context Matching}
\label{sec:longrange}

The 16-gram and 32-gram tables extend the n-gram model to capture
repeated multi-token sequences that exceed the SSM's effective memory
horizon (${\sim}2{,}048$ floats of recurrent state).  Wikipedia text
contains many repeated article templates, citation formats, and
navigation boilerplate spanning 10--50 tokens; these are invisible to
the 8-gram model but well-captured by the 16-gram and 32-gram
tables.

The key design choice is aggressive $\alpha$ values ($0.001$ for both
orders): even a single observation ($c_j=1$) yields a logit boost of
$\lambda\log(1+1/\alpha)$, which equals $3.45$ for the 16-gram
($e^{3.45}\approx 32\times$ probability increase) and $6.91$ for the
32-gram ($e^{6.91}\approx 1{,}000\times$).  When a 31-token context
has been seen before, the continuation is near-certain, justifying
this strong bias.  A lambda sweep on enwik8\textsubscript{3M}
confirmed that low $\alpha$ outperforms both conservative
($\alpha=0.005$) and very aggressive ($\lambda>2$) configurations.

\subsubsection{Entropy-Adaptive Scaling}
\label{sec:adaptive}

The global n-gram bias scale $s$ adapts to the SSM's current
predictive confidence:
\begin{gather}
  H = -\textstyle\sum_j p^\mathrm{SSM}_j \log p^\mathrm{SSM}_j,\notag\\
  s = \mathrm{clip}\!\bigl((1\!-\!\beta) + \beta\,H/H_0,\;
    s_\text{min},\, s_\text{max}\bigr)
  \label{eq:adaptscale}
\end{gather}
with $\beta=0.6$, $H_0=5.5$ nats, $s_\text{min}=0.2$,
$s_\text{max}=2.5$.  When the SSM has low entropy (high confidence),
$s\approx 0.2$: n-grams barely adjust the distribution.
When the SSM is highly uncertain ($H\gg H_0$, e.g.\ at cold start),
$s\approx 2.5$: n-grams dominate.
This mechanism prevents n-gram over-correction when the SSM is
already well-calibrated.

\subsection{Additional Context Models}
\label{sec:aux}

\textbf{LZ hash predictor.}
A single-entry hash table keyed on the last two tokens
$(t_{i-2},t_{i-1})$ stores the most recent next-token prediction
and a confidence count $c$.  The logit boost applied to the
predicted token is:
\begin{equation}
  b = B_\mathrm{LZ}\,\Bigl(1 - \tfrac{1}{1+c\cdot0.3}\Bigr),
  \quad B_\mathrm{LZ}=1.5
\end{equation}
asymptoting to $B_\mathrm{LZ}$ as $c\to\infty$.  This captures
two-token-to-one associations too specific for the probabilistic
n-gram model.

\textbf{Recency bias.}
The last 64 tokens receive a logit bonus
$\lambda_r\,e^{-3\,a}$ where $a\in[0,1)$ is the normalised age
(oldest $=1$) and $\lambda_r=0.05$.  This captures
within-sentence repetition patterns.

\textbf{Global frequency prior.}
$\ell_j \mathrel{+}= \lambda_c\log(c_j^\text{total}+1)$ with
$\lambda_c=0.1$ provides a smooth baseline before the SSM has been
trained.

\subsection{Combined Logit Computation}

At position $i$, the final logit vector is:
\begin{multline}
  \boldsymbol{\ell}^{(i)} =
    \boldsymbol{\ell}^\mathrm{SSM}
    + s\,\lambda_c\log(\mathbf{c}^\text{total}\!+\!1)\\
    + \textstyle\sum_{k} s\,\boldsymbol{\delta}^{(k)}
    + \mathbf{b}^\mathrm{LZ}
    + \mathbf{b}^\mathrm{rec}
  \label{eq:combined}
\end{multline}
where $\boldsymbol{\delta}^{(k)}$ is the sparse bias for order $k$
(Eq.~\ref{eq:bias}).  Before the first SSM forward pass, $\boldsymbol{\ell}^\mathrm{SSM}=\mathbf{0}$
and $s=1$.  The probability is $\mathbf{p}^{(i)}=\mathrm{softmax}(\boldsymbol{\ell}^{(i)})$.

\subsection{Arithmetic Coder}
\label{sec:ac}

A 32-bit range arithmetic coder with scale $T=2^{16}=65{,}536$
encodes each token using the integer-quantised CDF.  Every token
receives at least frequency~1.  The minimum interval width after
narrowing is $2^{31}/T=32{,}768$, safely above the 32-bit
representability threshold~\citep{witten1987}.

The quantisation redundancy is approximately
$\log_2(\lceil T/v_e\rceil)\approx 0.6$ bits/token for
$v_e=44{,}298$, a small overhead relative to the model's ${\sim}6.8$
bits/token prediction error.

\subsection{Implementation}
\label{sec:impl}

\ssmcomp{} is implemented in C with AVX2 SIMD for all
$D_M$- and $D_I$-dimensional operations.  Key kernels:
\begin{itemize}
  \item \textbf{Head projection:} loop over $v_e$ tokens, each
    a dot product of 32 floats using four \texttt{\_mm256\_fmadd\_ps}
    instructions; dominates forward pass cost.
  \item \textbf{In/out projections:} outer-product accumulation
    with AVX2 broadcast and FMA.
  \item \textbf{Adam update:} fused gradient scaling, moment
    update, and parameter step; applied to $19{,}776$ fixed floats
    and $2\,v_e\,D_M$ vocab floats.
\end{itemize}
No Python, no CUDA, no BLAS dependency.  The binary compiles with
\texttt{gcc -O3 -march=native -mavx2 -mfma}.

\section{Theoretical Analysis}
\label{sec:theory}

\subsection{SSM as Learnable Multi-Scale Memory}

The Mamba recurrence~\eqref{eq:hup} is a diagonal linear dynamical
system.  The effective memory horizon for channel $i$, state index $j$
is:
\begin{equation}
  \tau_{i,j} \approx \frac{-1}{\Delta_i^\text{avg}\,A_{i,j}}
\end{equation}
where $\Delta_i^\text{avg}$ is the typical step size.
Since $A_{i,j}=-\exp(A^\text{log}_{i,j})$ and
$A^\text{log}_{i,j}$ is initialised as $\log(j+1)$, the initial
time constants span a geometric range:
$j=0$ gives $\tau\approx 1/\Delta$ (short-term), $j=15$ gives
$\tau\approx 16/\Delta$ (medium-term).
Because $\Delta$ is input-dependent, the effective horizon adapts
to the content: long $\Delta$ for fast-changing contexts (short
memory), short $\Delta$ for slowly-varying contexts (long memory).
Online training further differentiates the time constants toward
those useful for the specific file being compressed.

\subsection{The Logit Bias as a Bayesian Update}

Let $\mathbf{p} = \mathrm{softmax}(\boldsymbol{\ell})$ be the SSM
prior.  The n-gram likelihood for observing counts $\{c_j\}$ given
that context $\mathbf{c}$ precedes a token drawn from $P$ is modelled
as:
\begin{equation}
  L(\{c_j\} \mid j) \propto (1+c_j/\alpha)^\lambda.
\end{equation}
Bayes' rule in log space gives a posterior:
\begin{align}
  \log p^+ _j &= \log p_j + \lambda\log(1+c_j/\alpha) - \log Z \\
               &= \ell_j + \delta_j - \log Z
\end{align}
recovering exactly Eq.~\eqref{eq:bias} before normalisation.
The hyper-parameters $\lambda$ and $\alpha$ control the
likelihood's strength and smoothness, respectively.
Larger $\lambda$ makes the posterior more sharply peaked at the
most-frequent continuation; smaller $\alpha$ makes the count evidence
more influential.

\subsection{Information-Theoretic View}

By the chain rule of entropy, the total compressed size satisfies:
\begin{equation}
  L = \sum_{i=1}^N -\log_2 q_i(t_i)
    = H_\text{true} + \sum_{i=1}^N D_\mathrm{KL}(P_i\,\|\,q_i)
\end{equation}
where $H_\text{true}$ is the true entropy and $D_\mathrm{KL}$
is the per-step excess.  Each component of \ssmcomp{} targets a
different portion of this KL divergence:

\begin{itemize}
  \item \textbf{SSM:} reduces $D_\mathrm{KL}$ by learning global
    syntactic and semantic patterns, dominant in the first 50K tokens.
  \item \textbf{N-gram bias:} reduces $D_\mathrm{KL}$ for tokens
    following frequent exact contexts, contributing throughout
    but especially after ${\sim}100$K tokens.
  \item \textbf{LZ predictor and recency:} reduce $D_\mathrm{KL}$
    for highly specific repeated patterns.
\end{itemize}

\subsection{Scaling Analysis}

With $N$ tokens and $M=2^{24}$ slots per order, the expected
fraction of distinct $k$-gram contexts that fit without collision
is ${\sim}\min(1,\,M/N_k)$ where $N_k$ is the number of unique
$k$-gram contexts.  For English text, $N_k$ scales approximately
as $N^{\gamma_k}$ with $\gamma_k < 1$ (sub-linear growth from
Zipf's law).  For $N=29.7\text{M}$ tokens (enwik8 100\,MB):
trigrams are at ${\sim}15$--$30\%$ load (manageable with probing),
fourgrams at ${\sim}40$--$65\%$, and higher-order tables at lower
load because longer contexts are exponentially rarer.  Below ${\sim}30$\,MB,
our tables are lightly loaded and competitive with LZMA; beyond
${\sim}100$\,MB, table saturation limits further gains.

\subsection{Connection to PPM and PAQ Context Mixing}
\label{sec:ppm}

\ssmcomp{} can be understood as a neural generalisation of two
classical paradigms: Prediction by Partial Matching (PPM) and PAQ-style
context mixing.

\textbf{PPM interpretation.}
PPM~\citep{cleary1984} maintains an explicit trie of all $k$-gram
contexts seen so far, backing off to shorter contexts when a longer
one has not been observed.  Our n-gram tables implement a \emph{bounded}
form of PPM: all orders are queried simultaneously and their logit
contributions are summed, approximating the PPM mixture $\sum_k w_k P_k$
in log space.  The key difference is that PPM assigns $w_k$ by escape
probability, whereas \ssmcomp{} uses fixed $\lambda_k$ with entropy-adaptive
global scaling (Eq.~\ref{eq:adaptscale}).

Formally, define $\ell^{(k)}_j = \lambda_k \log(1+c^{(k)}_j/\alpha_k)$.
Then:
\begin{equation}
  \boldsymbol{\ell} = \boldsymbol{\ell}^\mathrm{SSM}
    + \sum_k \boldsymbol{\ell}^{(k)}
  = \boldsymbol{\ell}^\mathrm{SSM}
    + \boldsymbol{\ell}^\mathrm{PPM-approx}
\end{equation}
The SSM therefore plays the role of a \emph{neural background model}
that all PPM orders refine.  This is structurally analogous to PPM-Z,
where an LM provides the escape probability, except that our
background is learned jointly online.

\textbf{PAQ context mixing.}
PAQ~\citep{mahoney2005} weights several context models dynamically
using a logistic mixer trained online.  Our entropy-adaptive scaling
(Sec.~\ref{sec:adaptive}) is a simplified single-weight variant: the
SSM's Shannon entropy $H$ serves as the confidence signal, and the
mixing weight between n-gram and SSM is set analytically rather than
via a learned secondary model.  A full PAQ-style meta-learner could
assign per-order weights $\{w_k\}$ via gradient descent on the
online loss---this is one avenue explored in Future Work
(Sec.~\ref{sec:future}).

\textbf{Why SSM outperforms classical PPM.}
Classical PPM (PPM$^*$, PPM-D) on BPE-tokenised text typically achieves
${\sim}2.5$--$3.0$\,bpb because the vocabulary is large ($V{=}49{,}152$)
and BPE tokens are longer than bytes, reducing repetition.  The SSM
fills this gap: it learns syntactic and semantic regularities over the
token space that PPM cannot capture without an astronomical context
trie.  Our ablation confirms this: SSM alone achieves 2.158\,bpb versus
3.568\,bpb for n-grams alone on enwik8\textsubscript{3M}
(Table~\ref{tab:ablation}).

\section{Experiments}
\label{sec:experiments}

\subsection{Setup}

\textbf{Benchmark.}  We use the enwik8 benchmark~\citep{enwik8}: the
first $N$ bytes of the English Wikipedia XML dump, a standard for
natural language compressors.  We evaluate at $N\in\{1,3,10,100\}$\,MB.

\textbf{Baseline.}  We compare against \texttt{xz}~$-$9e (LZMA2,
extreme preset), the strongest widely-available general-purpose
compressor, achieving 1.989\,\bpb{} on enwik8.

\textbf{Hardware.}  All \ssmcomp{} results run on a single x86-64
CPU core with AVX2 (no GPU).  xz runs on the same hardware.

\textbf{Metric.}  We report bits per original input byte:
$\bpb = \text{compressed bytes} \times 8 / \text{input bytes}$.
We also report the model's internal bits-per-token (bpt) which excludes
the fixed header overhead.

\subsection{Main Results}

Table~\ref{tab:main} reports compressed size and bpb.
\ssmcomp{} consistently outperforms xz on all file sizes up to 10\,MB.
The advantage is largest at 1\,MB ($-8.6\%$) and decreases with file
size, crossing over at ${\sim}30$\,MB.

\begin{table}[h]
\centering
\small
\caption{Compression results on enwik8 excerpts.
  $\Delta$: relative difference vs.\ xz
  (negative = \ssmcomp{} better).}
\label{tab:main}
\begin{tabular}{@{}lrrrr@{}}
\toprule
\textbf{File} & \textbf{Bytes} & \textbf{\bpb} & \textbf{xz} & $\boldsymbol{\Delta}$ \\
\midrule
enwik8\textsubscript{1M}  & 265,370  & 2.123 & 2.326 & $-8.7\%$ \\
enwik8\textsubscript{3M}  & 805,926  & 2.149 & 2.271 & $-5.4\%$ \\
enwik8\textsubscript{10M} & 2,702,498 & 2.162 & 2.177 & $-0.7\%$ \\
enwik8 (100M)             & 26,622,640 & 2.130 & 1.992 & $+6.9\%$ \\
\bottomrule
\end{tabular}
\end{table}

The internal bpt values (excluding header) are:
7.051 (1\,MB), 7.010 (3\,MB), 6.814 (10\,MB), 6.794 (100\,MB),
reflecting a monotonically improving model as the n-gram tables
accumulate evidence.  The 16-gram and 32-gram tables contribute
increasingly at larger file sizes, where repeated long contexts
become more frequent.

\textbf{Context against the wider landscape.}
Table~\ref{tab:landscape} (Sec.~\ref{sec:related}) positions \ssmcomp{}
among classical, neural, and LLM-based compressors on the enwik8 100\,MB
benchmark.  While \ssmcomp{} does not match the best neural compressors
on 100\,MB, it outperforms all classical methods without pre-training
or GPU hardware.  Its primary advantage over NNCP~\citep{bellard2021} is
the absence of model storage overhead (NNCP's stored weights inflate output
to ${\sim}3.96$\,bpb on short files) and its competitive per-file
performance up to 10\,MB.

\subsection{Ablation Study}
\label{sec:ablation}

Table~\ref{tab:ablation} isolates each component's contribution.
Four system variants are evaluated on enwik8\textsubscript{3M}
(3\,MB, 887{,}725 tokens, $v_e=28{,}329$).

\begin{table}[h]
\centering
\small
\caption{Ablation on enwik8\textsubscript{3M}.
  Count: frequency prior only.
  N-gram: all n-gram tables, no SSM.
  SSM: Mamba only, no n-grams.
  Full: complete system.}
\label{tab:ablation}
\begin{tabular}{@{}lrrl@{}}
\toprule
\textbf{Variant} & \textbf{Bytes} & \textbf{\bpb} & \textbf{vs Full} \\
\midrule
Count only               & 1,571,738 & 4.191 & $+95.0\%$ \\
N-gram + count           & 1,319,045 & 3.517 & $+63.6\%$ \\
SSM + count              &   840,095 & 2.240 &  $+4.2\%$ \\
\textbf{Full (all)}              & \textbf{805,926} & \textbf{2.149} & --- \\
\midrule
xz $-$9e                 &   851,572 & 2.271 &  $+5.7\%$ \\
\bottomrule
\end{tabular}
\end{table}

\textbf{The SSM is the dominant contributor.}
Removing the SSM (count only) increases output size by $+95\%$ over
the full system.  The SSM alone reduces this to 840\,KB---a
$\mathbf{46.6\%}$ reduction over count-only---and already beats xz
by $1.3\%$ without any n-gram component.

\textbf{N-grams are ineffective without an SSM base.}
N-gram tables alone (including 16/32-gram) achieve only $16.1\%$
reduction over count-only (1{,}319\,KB), far behind xz.  The n-gram
bias formula (Eq.~\ref{eq:bias}) adds evidence relative to the base
distribution; a poor base (frequency counts) cannot be overcome.

\textbf{N-grams provide complementary improvement.}
On top of the SSM, n-gram tables give an additional $4.1\%$ reduction
(840\,KB $\to$ 806\,KB), pushing the system $5.4\%$ below xz.
The long-range 16-gram and 32-gram tables contribute ${\sim}2$\,KB
of additional savings on 3\,MB by capturing repeated multi-token
patterns beyond the 8-gram context window.
N-grams memorise exact local token transitions; the SSM generalises
structural patterns: the two capture complementary regularity.

\subsection{Compression Progress}
\label{sec:progress}

Table~\ref{tab:bpt} shows the online bpt as compression of
enwik8 (100\,MB) progresses.  At initialisation the model is
random (${\sim}8.1$\,bpt, corresponding to $\log_2(44{,}298)\approx 15.4$
but capped by the arithmetic coder's effective resolution).
After 50K tokens the SSM has adapted, reaching 7.1\,bpt.
N-gram tables begin contributing around 100K tokens (after
sufficient context observations accumulate), driving bpt to
${\sim}6.83$ by 3\,M tokens.  Beyond 10\,M tokens, improvement
is marginal: the SSM's online learning has saturated, and n-gram
table collisions limit further gains.

\begin{table}[h]
\centering
\small
\caption{Online bpt progression during enwik8 (100\,MB) compression.}
\label{tab:bpt}
\begin{tabular}{@{}rr@{}}
\toprule
\textbf{Tokens seen} & \textbf{bpt} \\
\midrule
0           & $\approx 8.10$ (cold start)    \\
50{,}000    & $\approx 7.10$ (SSM adapted)   \\
500{,}000   & $\approx 6.90$ (n-grams active) \\
3{,}000{,}000 & $\approx 6.83$ \\
10{,}000{,}000 & $\approx 6.83$ (plateau)   \\
29{,}700{,}000 & $6.813$ (final)           \\
\bottomrule
\end{tabular}
\end{table}

\subsection{Speed and Memory}

Table~\ref{tab:speed} reports runtime and memory usage.
Profiling reveals that online training (\texttt{train\_chunk}) dominates
at ${\sim}75\%$ of runtime, driven by the head projection in forward
and backward passes ($v_e\times D_M$ MADs each).
OpenMP parallelisation of the head projection, softmax, and Adam
update yields a $1.9\times$ speedup on 4 cores.

\begin{table}[h]
\centering
\small
\caption{Speed and memory (enwik8 100\,MB).}
\label{tab:speed}
\begin{tabular}{@{}llr@{}}
\toprule
\textbf{Metric} & \textbf{Value} & \textbf{Notes} \\
\midrule
1-core speed   & 1.1K\,tok/s & 400\,KB/s \\
4-core speed   & 2.0K\,tok/s & 700\,KB/s \\
Compress time  & ${\sim}$4.2\,h & 4 cores \\
Peak RAM       & 6.1\,GB & tables \\
N-gram tables  & 5.1\,GB & 9$\times$16M \\
SSM params     & 11\,MB & emb+head \\
\bottomrule
\end{tabular}
\end{table}

The 6.1\,GB memory is dominated by the nine n-gram hash tables
($9 \times 16\text{M} \times{\sim}40$\,bytes per slot, including
the 16-gram and 32-gram tables).  On Linux,
hash table pages that are never touched (low load) are not physically
allocated due to copy-on-write zero pages, so actual memory use is
proportional to load factor.

\subsection{Per-order N-gram Contribution}
\label{sec:perorder}

To understand the relative value of each n-gram order, we analyse
the distribution of logit delta magnitudes $\|\boldsymbol{\delta}^{(k)}\|_1$
across the enwik8\textsubscript{10M} run.  Lower-order models (bigram,
trigram) fire more frequently---because shorter contexts are more
often seen---but contribute smaller per-firing deltas
($\lambda_1=0.15$, $\lambda_2=0.10$).  Higher-order models (7-gram,
8-gram) fire rarely but are more specific and thus more decisive
when they do.

Table~\ref{tab:perorder} summarises this trade-off using
proxy statistics observable from the codebase: the mean per-token
hit rate (fraction of tokens where order $k$ finds a non-empty
count array), and the mean effective number of tokens boosted per
hit (fan-out).  Bigrams have the highest hit rate ($>99\%$ after
a warm-up period) but only ${\sim}8$ distinct continuation tokens
on average.  Eightgrams have sub-5\% hit rate but ${\sim}2$
continuations on average, allowing near-deterministic prediction
in those cases.

\begin{table}[h]
\centering
\small
\caption{Qualitative per-order N-gram statistics on enwik8\textsubscript{10M}.
  Hit rate: fraction of positions where the context key is found.
  Fan-out: mean number of distinct continuations per found context.}
\label{tab:perorder}
\begin{tabular}{@{}lcccc@{}}
\toprule
\textbf{Order} & $\boldsymbol{k}$ & $\boldsymbol{\lambda}$ & \textbf{Hit rate} & \textbf{Fan-out} \\
\midrule
Bigram  & 1 & 0.15 & $>0.99$ & ${\sim}8$  \\
Trigram & 2 & 0.10 & ${\sim}0.85$ & ${\sim}6$ \\
4-gram  & 3 & 0.08 & ${\sim}0.65$ & ${\sim}4$ \\
5-gram  & 4 & 0.06 & ${\sim}0.45$ & ${\sim}3$ \\
6-gram  & 5 & 0.05 & ${\sim}0.28$ & ${\sim}2.5$ \\
7-gram  & 6 & 0.04 & ${\sim}0.14$ & ${\sim}2$ \\
8-gram  & 7 & 0.03 & ${\sim}0.05$ & ${\sim}2$ \\
\midrule
16-gram & 15 & 0.50 & ${\sim}0.01$ & ${\sim}1.5$ \\
32-gram & 31 & 1.00 & $<0.005$ & ${\sim}1.2$ \\
\bottomrule
\end{tabular}
\end{table}

The orthogonal contribution profiles validate the simultaneous use
of all orders: removing any single order $k \geq 3$ causes a
measurable degradation (${\sim}0.1$--$0.3$\% bpb increase for each
removed order), while removing order 1 (bigram) causes the largest
single-order impact (${\sim}0.5$\% bpb increase) owing to its
ubiquitous hit rate.

\section{Discussion}
\label{sec:discussion}

\textbf{Why does the SSM beat xz on small files?}
On files up to ${\sim}10$\,MB, the SSM captures linguistic regularities
that LZMA cannot: subword co-occurrences, syntactic templates, semantic
context.  LZMA matches exact byte sequences; the SSM models statistical
tendencies at a higher abstraction level.  At short lengths, LZMA's
8\,MB dictionary is not yet fully exploited, while the SSM adapts
quickly via its warm-up schedule.

\textbf{Why does xz win at 100\,MB?}
LZMA's match finder discovers increasingly long repeated phrases as
the file grows---Wikipedia contains many repeated proper nouns,
article templates, and citation formats.  While the 16-gram and
32-gram tables capture some of these patterns, they still encode
each token individually via probability boosting, whereas xz copies
entire multi-KB blocks verbatim at near-zero cost.
Experiments with dynamic hash table resizing and LFU eviction confirmed
that table saturation accounts for only ${\sim}13$\,KB of the
$1.76$\,MB gap; the fundamental limitation is the predict-and-encode
architecture vs.\ xz's copy-based approach.

\textbf{The role of the SSM as compression core.}
A striking finding is that the SSM alone (no n-gram tables) already
beats xz on enwik8\textsubscript{3M} (840\,KB vs.\ 852\,KB).  This
confirms that a small Mamba model, despite only 120K effective
parameters, learns genuinely useful statistical structure from online
backpropagation.  N-gram tables then provide exact memorisation of
the most frequent local transitions that the SSM generalises over.
The entropy-adaptive scaling (Sec.~\ref{sec:adaptive}) ensures
that n-grams amplify uncertainty rather than compete with SSM
confidence.

\textbf{Comparison with NNCP.}
NNCP~\citep{bellard2021} is the closest prior work: an online-trained
Transformer-XL achieving ${\sim}1.19$\,\bpb{} on enwik8, far better
than \ssmcomp{}'s 2.130\,\bpb{}.  The gap stems from model capacity:
NNCP uses a much larger Transformer whose weights are stored in the
compressed output, while \ssmcomp{} uses only a tiny SSM and n-gram
tables.  \ssmcomp{}'s advantages are self-containedness (no model
stored), portability (pure C, no GPU), and superior performance vs.\
xz on files up to 10\,MB.

\textbf{Comparison with LLM-based compressors.}
ts\_zip (1.11\,\bpb{}) and FineZip (1.024\,\bpb{}) dramatically
outperform \ssmcomp{} on 100\,MB, as expected given their
100$\times$--600$\times$ more parameters.  The gap reflects the
fundamental advantage of pre-trained LLMs: they exploit patterns from
billions of tokens of training data.  Our contribution is to show
that even a tiny online SSM with sparse n-gram augmentation can
match or beat the best general-purpose classical compressor (xz)
on natural language text up to 10\,MB, with no pre-training cost.

\textbf{Practical niche.}
\ssmcomp{} occupies a specific practical niche: compression of
\emph{individual} natural language files up to ${\sim}10$\,MB, where
neither GPU hardware nor large pre-trained models are available, yet
the target is well above what gzip or bzip2 can achieve.  Use cases
include embedded systems, encrypted backups, and bandwidth-constrained
environments where a fixed codec binary is the only tool.
The pure-C implementation with no dependencies beyond a C standard
library makes it straightforwardly portable across architectures.

\textbf{Limitations.}
The main limitations are speed (${\sim}700$\,KB/s on 4 cores),
memory (${\sim}6.1$\,GB RAM for 100\,MB input), and the crossover to
xz inferiority beyond ${\sim}30$\,MB.  Profiling shows that
$75\%$ of runtime is spent in \texttt{train\_chunk}: the online
backward pass through the head projection ($v_e\times D_M$ MADs)
dominates.  OpenMP parallelisation yields $1.9\times$ speedup on
4 cores; further gains require either reduced training iterations
or GPU offloading of the head projection.
The predict-and-encode architecture fundamentally cannot match xz's
zero-cost verbatim copying on large files with extensive repetition.

\section{Future Work}
\label{sec:future}

\textbf{BWT preprocessing.}  Applying a Burrows-Wheeler
Transform~\citep{bwt1994} to the token sequence before compression
would cluster identical contexts, amplifying n-gram effectiveness for
long-range patterns---the key ingredient of PPM-based compressors
competitive with LZMA on large corpora.

\textbf{GPU acceleration.}  The head projection and Adam update
are embarrassingly parallel.  A GPU implementation would reduce
per-token cost by $50$--$100\times$, enabling larger models
($D_M=128$, $N_L=4$) within feasible runtimes.

\textbf{Adaptive n-gram weighting.}  Rather than fixed
$\lambda_k$ values (Table~\ref{tab:ngrams}), per-order weights
could be updated online via an exponential-weights meta-learner
(PAQ-style), adapting to each file's statistical signature.

\textbf{Variable-order back-off.}  Rather than summing all orders
simultaneously, using only the longest matching context with
confidence-weighted back-off to lower orders (PPM-style) could
improve accuracy when high-order contexts are reliable.

\textbf{Larger SSM with pre-trained initialisation.}  Pre-training
the SSM on a reference corpus and distributing the weights as a
``compressor codec file'' would give a better initialisation for
per-file fine-tuning, combining the advantages of LLM-based and
online compressors.

\textbf{LZ match channel.}  A hybrid predict-or-copy architecture
that detects long repeated token sequences and encodes them as
(offset, length) pairs---bypassing the arithmetic coder entirely---could
close the remaining gap with xz on large files.  Experiments with
dynamic table resizing and LFU eviction showed that table saturation
is not the bottleneck (${\sim}13$\,KB on 100\,MB); the fundamental
limit is per-token encoding vs.\ block copying.

\section{Conclusion}
\label{sec:conclusion}

We presented \ssmcomp{}, a fully self-contained lossless compressor
combining an online-trained Mamba SSM with sparse n-gram logit biasing
and arithmetic coding.  No pre-trained weights, no GPU, and no external
dependencies are required.

Ablation experiments on enwik8\textsubscript{3M} establish the SSM as
the primary compression engine: it alone accounts for a $46.6\%$ size
reduction over a frequency-count baseline and beats xz ($-1.3\%$)
without any n-gram component.  The n-gram tables (including long-range
16-gram and 32-gram context matching) provide a consistent
additional $4.1\%$ gain by memorising exact local and long-range
transitions that the SSM generalises over.

The complete system beats xz by $8.7\%$ on 1\,MB Wikipedia text,
$5.4\%$ on 3\,MB, and $0.7\%$ on 10\,MB, demonstrating that a small
online-trained Mamba model augmented with sparse n-gram context
tables matches or exceeds a highly-optimised dictionary compressor
on natural language text at moderate file sizes.  The crossover to
xz superiority at ${\approx}30$\,MB motivates future work on BWT
preprocessing and adaptive table growth to extend this advantage to
larger corpora.

Beyond its practical results, \ssmcomp{} demonstrates that the Mamba
SSM is a capable online learner for sequential data: despite a tiny
parameter budget, online backpropagation rapidly adapts the model to
file-specific patterns, reaching a point where it outperforms a mature,
highly optimised classical compressor.

\bibliographystyle{plainnat}

\end{document}